# 📞👨⛵🐳👌; or “Call me Ishmael” – How do you translate emoji?

**Will Radford** **Andrew Chisholm** **Ben Hachey** **Bo Han**
Hugo.ai
58-62 Kippax Street
Surry Hills
NSW 2010
(wradford|achisholm|bhachey|bhan)@hugo.ai

## Abstract

We report on an exploratory analysis of Emoji Dick, a project that leverages crowdsourcing to translate Melville’s Moby Dick into emoji. This distinctive use of emoji removes textual context, and leads to a varying translation quality. In this paper, we use statistical word alignment and part-of-speech tagging to explore how people use emoji. Despite these simple methods, we observed differences in token and part-of-speech distributions. Experiments also suggest that semantics are preserved in the translation, and repetition is more common in emoji.

## 1 Introduction

Emoji are pictographic characters commonly used in informal written communication, originally in SMS messages, but increasingly in all social media. Emoji represents a range of faces, animals, weather, emotions and activities, compensating for a lack of verbal and non-verbal cues in face-to-face communication (Davis and Edberg, 2016; Dresner and Herring, 2010). Widespread support for emoji input on desktops and mobile devices has led to a rapid adoption of emoji as a communication tool and a part of popular culture, spawning a *Word of the Year* in 2015 (Oxford Dictionaries, 2015) and, somewhat dubiously, a feature film slated for a 2017 release (IMDb, 2016). Davis and Edberg (2016) also suggest that the limited coverage and a lack of formal grammar in emoji lead to multiple ambiguous messages, and point out the playful aspect of composing and decoding emoji. At its tersest, this can result in a film plot encoded into a single tweet (MacLachlan, 2016): “🌏💧. . . 🐛. . . 👑#dune”. Despite heavy constraints of its form, emoji are used in some surprising linguistic contexts that merit further study.

| TEXT | EMOJI |
|---|---|
| MOBY DICK; OR THE WHALE | 🐳 |
| CHAPTER 1. | 👍1️⃣➿👍👍 |
| Loomings. | 👀👇👎☁☁ |
| Call me Ishmael. | 📞👨⛵🐳👌 |

Figure 1: The title and first lines of Emoji Dick.

Most analysis of emoji considers its usual context amongst regular text, focusing on geographic preferences (Swiftkey, 2015; Ljubešić and Fišer, 2016), single-word translations (Dimson, 2015) and sentiment (Kralj Novak et al., 2015). Other work has focused on emoticons (e.g. “:)”), uncovering stylistic variation (Schnoebelen, 2012) and demographic factors associated with usage (Hovy et al., 2015). Different types of ambiguity have also been identified: subjective interpretations of symbols, as well as different renderings of what should be the same symbol (Miller et al., 2016). Mitchell et al. (2015) observe that emoji are less likely to occur in the language of schizophrenics, possibly related to a “flat affect” symptom of the mental disorder.

This paper analyses a specific instance where emoji have been used in a context usually reserved for fully-fledged languages. Emoji Dick (Benenson, 2010) is a fascinating project that used Amazon Mechanical Turk to translate the novel “Moby-Dick; or, The Whale” (Melville, 1851) into emoji (e.g. Figure 1). Three turkers were paid $0.05 to translate each sentence, and a second round of turkers paid $0.02 to vote for the best translation. Funded entirely by donations through the crowdsourcing platform Kickstarter, soft- and hard-bound copies are available for sale. The idiosyncratic nature of the task raises several key questions: how does one translate a work from the English literary canon into emojis? And, what can statistical analysis techniques from natural language processing tell us about it?

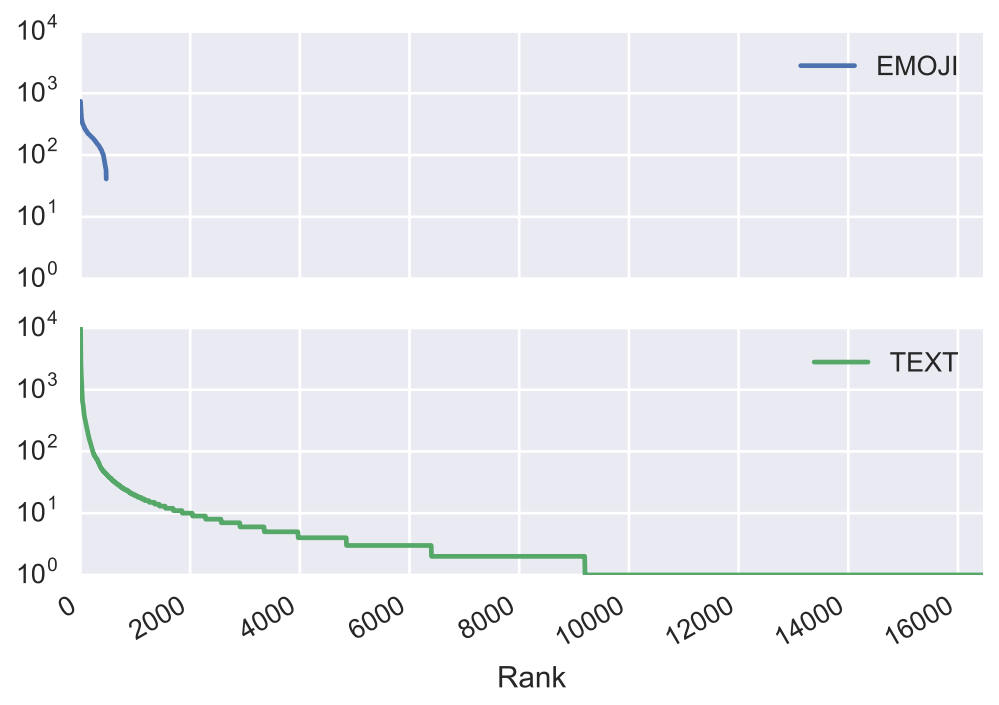

Figure 2: Log-counts of emoji and tokens at different ranks.

## 2 Corpus Analysis

The dataset contains 9,971 pairs of text and emoji translations. We tokenise the text using the default tokeniser from NLTK (Bird et al., 2009). Digits and stopwords are removed and all tokens are lowercased.

The descriptive statistics of the two corpora, TEXT and EMOJI, show a few differences. TEXT has a larger and sparser vocabulary than EMOJI when constructing sentences. There are 207,194 text tokens (16,454 types) and 93,342 emoji tokens (470 types). Furthermore, emoji sentences are roughly half the length of textual sentences with respective means of 9.4 to 20.8. Figure 2 plots the log-counts of two corpora token or emoji frequencies against their ranks. As is common in Zipfian distributions, the TEXT frequencies quickly reduce with rank, but EMOJI has a heavier tail and no approach to low frequencies.

Table 1 shows the 20 most common tokens and emoji. Unsurprisingly, "whale" is the most common non-stopword token, while other thematic tokens (e.g. "man", "ship", "sea", "boat") and character names (e.g. "ahab", "captain"') are also common. The three most common emoji reference characters: people (e.g. 👲💁👸), and, of course, 🐳. Less obvious emoji include references to objects (e.g. ☕🚑🎓🔑) and places (e.g. 🏪), though not 🚢 or 🚣 as found in the common tokens. Symbols (e.g. ✖❔❗) are also present, as are other non-facial body parts (e.g. 👀👏).

## 3 Aligning emoji to tokens

The analysis above makes no attempt to characterise how TEXT was translated to EMOJI. To explore this, we learn an IBM translation model one (Brown et al., 1993) using NLTK's implementation with default parameters. We prepare the parallel text by removing English stopwords and punctuation and filtering pairs missing input or output, leaving 9,734 pairs. We opt for the simplest model with the fewest assumptions about the "language pair", but with the caveat that it does not account well for token phrases aligning to a single emoji, as a phrase-based translation model might allow. We did explore some of the more advanced IBM models and sequence-to-sequence neural models (Sutskever et al., 2014), but these did not seem to yield readily-interpretable results. Recall that despite turkers vote for the best translations, the translations are expected to be extremely noisy, due to the high subjectivity of human emoji comprehension, and the scarcity of native speakers.

| Token | $n$ | Emoji | $n$ |
|---|---|---|---|
| whale | 1029 | 👲 | 743 |
| one | 898 | 💁 | 724 |
| like | 572 | 👸 | 669 |
| upon | 561 | ☕ | 637 |
| ahab | 511 | 🚑 | 626 |
| man | 497 | 🎓 | 607 |
| ship | 464 | 🔑 | 598 |
| old | 435 | 🎡 | 574 |
| ye | 433 | 🏪 | 556 |
| would | 429 | ✖ | 537 |
| though | 380 | ❔ | 511 |
| sea | 367 | 🐳 | 496 |
| yet | 344 | ❕ | 442 |
| time | 325 | 👀 | 439 |
| captain | 323 | 👦 | 438 |
| long | 315 | 🙎 | 419 |
| still | 312 | 👏 | 415 |
| said | 299 | ❗ | 407 |
| great | 288 | 😳 | 399 |
| boat | 286 | 👨 | 379 |

Table 1: The top 20 tokens and emoji by individual frequency.

Having trained the model over 100 iterations, we examine the translation table and find the most probable alignment from token to emoji. Table 2 shows the top-20 emoji ranked by their strongest individual alignment with the tokens that they align to. The most commonly-aligned emoji, 🐳 aligns to "whale", "sperm" and "whales", as does 🐬. Rather less satisfying is the strong alignments of "whale" to ♍, and $\epsilon$ where a token aligned to nothing, showing the extent of the noise and the tendency of the model to align away tokens. Plural forms align well, as do synonyms (e.g. "boat", "ship" and "boats" align to 🚤). Other alignments cover formal elements such as chapter headings (📖) and characters (👮👨), natural phenomena (☀️💧💀), actions (👀💤🖐) and sentiments (👍). Overall, most high-probability alignments seem reasonable, suggesting some consistency between translations.

| $e$ | $t$ | $p$ | $t$ | $p$ | $t$ | $p$ | $e$ | $t$ | $p$ | $t$ | $p$ | $t$ | $p$ |
|---|---|---|---|---|---|---|---|---|---|---|---|---|---|
| 🐳 | whale | .23 | sperm | .06 | whales | .05 | 👨 | man | .07 | men | .05 | ahab | .02 |
| 📔 | chapter | .15 | book | .05 | read | .03 | 🚢 | ship | .07 | lay | .02 | vessel | .02 |
| $\epsilon$ | one | .11 | whale | .11 | like | .06 | 💤 | bed | .06 | good | .03 | sleeping | .02 |
| 1️⃣ | chapter | .10 | two | .02 | one | .02 | 🌊 | whale | .06 | among | .03 | sperm | .03 |
| 👮 | captain | .08 | sir | .06 | know | .02 | 👍 | good | .06 | things | .03 | man | .03 |
| ☀️ | sun | .08 | sunrise | .03 | air | .03 | 💧 | water | .06 | sea | .05 | round | .02 |
| 6️⃣ | chapter | .08 | straight | .02 | beheld | .01 | 💀 | death | .05 | life | .03 | dead | .03 |
| 👀 | see | .08 | eyes | .05 | seen | .03 | ♍ | whales | .05 | sperm | .03 | ancient | .02 |
| 🚤 | boat | .07 | ship | .03 | boats | .03 | ❕ | oh | .05 | aye | .02 | old | .02 |
| 🖐 | hand | .07 | say | .03 | stop | .03 | 🌆 | whales | .05 | starbuck | .02 | fossil | .02 |

Table 2: The top-20 emoji by alignment probability ($p(e|t)$), and the top 3 tokens they align to. We also list $\epsilon$ as some tokens (i.e. "one, whale, like") *align away* with relatively high probability.

## 4 Emoji parts-of-speech

We apply the BookNLP pipeline (Bamman et al., 2014) to the original text, which predicts part-of-speech (POS) tags, character clusters and many other linguistic metadata. Again, we expect this to be noisy, as the models are not adapted to 19th century literature. Nonetheless, we are able to create a distribution of noisy POS tags[1] for each token, then induce a distribution of POS for each emoji, by applying the chain rule and alignment probabilities above. Most emoji are NN, with the highest probability 🐳. Only 5 emoji have a different majority POS prefix – VB – including 👀 (i.e. "see" or "seen" from Table 2). Although the low probabilities and compounded noise of the model mean the results are hardly compelling, it is interesting that some emoji seem to align more strongly to verbs, suggesting their consistent use as actions. This preponderence of NN and VB alignments supports the observation made elsewhere that emoji have features in common with pidgin languages (Rosefield, 2014; Stockton, 2015).

Accumulating the POS probabilities for actual occurrence of tokens or emojis, we are able to characterise the whole distribution. Figure 3 shows that emoji concentrate mass into fewer categories: NN, JJ, RB and VB.

| Emoji | $p(VB|e)$ | Emoji | $p(NN|e)$ |
|---|---|---|---|
| 👀 | .37 | 🐳 | .70 |
| 🆚 | .36 | 🍃 | .56 |
| 🐺 | .30 | 💎 | .55 |
| 🍟 | .29 | 🆙 | .53 |
| 🎤 | .29 | 👮 | .52 |

Table 3: The top-5 emoji for VB and NN POS.

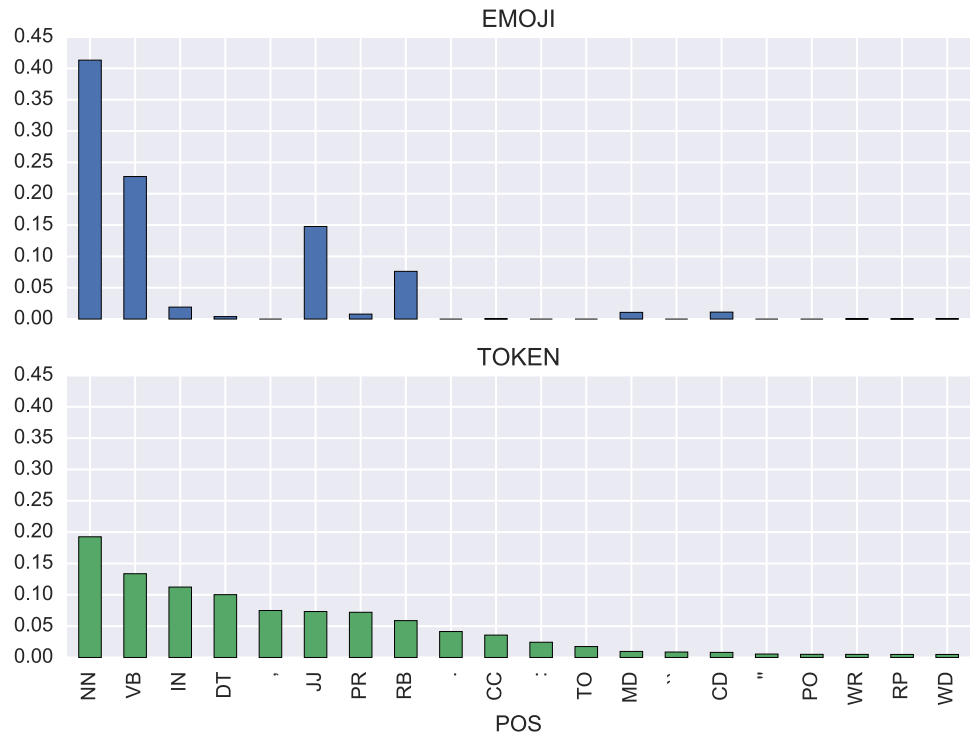

Figure 3: POS probability distributions.

## 5 Modelling repeats in emoji

As emoji are often used to convey extra layers of emotion or context, we examine how emoji are repeated. Counting common bigrams of emoji, the most frequently emoji repeat-bigrams are 💏💏 followed by 💪💪, ❕❕ and 👨👨. The probability of repeat-bigrams is much higher in emoji at 3.2% as opposed to 0.4% in text, indicating that there is a systematic difference in usage, although it is not clear whether they are used as intensifiers ("very strong") or counting mechanisms ("many boys").

[1] We attempt to reduce noise by considering the only the first two characters of the POS tag. NNP and NNS as NN.

## 6 Discussion

As intriguing as the dataset is, there are substantial challenges to working with it and drawing strong conclusions. Firstly, the setting is idiosyncratic and unlike standard current emoji usage, as it lacks the textual context in which emoji are usually found. Thus, findings in these contexts may not necessarily hold when applied to emoji found in informal discourse on the web. Indeed, a promising direction seems to be to learn embedding models that allow an elegant fusion of textual and emoji content (Barbieri et al., 2016; Eisner et al., 2016). The constraints of the limited vocabulary foster a creativity that, combined with the desire of the translator to produce a witty translation, only serves to increase the subjectivity of the dataset. This is exacerbated by the small size of the dataset, and while there are ordinal constraints in emoji (Schnoebelen, 2012), whether they hold for longer sequences is not clear.

The next issue to examine in future work is the failure of phrase-based translation models in this context, as this might uncover more interesting translations, as well as shed light on strange alignments (e.g. "whale" to $\epsilon$). Data collection is also important, as lack of data could prohibit more sophisticated models. Moreover, collecting a dataset in a more natural context, such as chat or social message data, might lead to more immediately-applicable conclusions. Exploring whether there is strong evidence for a consistent interpretation of emoji repeats using analysis of the textual corpus would be interesting. Multi-lingual differences in emoji usage are compelling – do certain languages exhibit features that correlate with different emoji usage? Moreover, emoji may be a productive pivot upon which to align web-scale bitext, which may drive other semantic resources such as paraphrase databases (Ganitkevitch et al., 2013). The overarching direction of this work are to characterise broader temporal trends about how people use emoji, how a community of users settles on specific interpretations of emoji and, perhaps, uncover more evidence for or against the hypothesis that emoji is a pidgin.

## 7 Conclusion

This paper presents a pilot study on how emoji are used in an unusual setting – verbatim translation. We use NLP techniques to compare the statistical properties of the two corpora (EMOJI and TEXT), showing that emoji operate in a far smaller space. Statistical alignment models from machine translation allow us to explore how translators mapped emoji onto tokens, including plural and synonymous forms. We also considered a noisily-induced POS distribution, showing that most emoji are nouns, with a few operating most frequently as verbs, and that the emoji POS distribution places weights on nouns and verbs, a phenomenon that bears some similarity to pidgins. Finally, we briefly explore how emoji are repeated, finding a much higher rate of repetition in emoji than text. Emoji offer a rare opportunity to study a rapidly evolving and increasingly important mode of communication that is complementary to text and speech and has parallels to human languages like pidgins.

## Acknowledgments

Many thanks to Fred Benenson and Chris Mulligan for having the wit to create the project and share the data 🙇🗿👦. Thanks also to the reviewers and readers at Hugo.ai for their insightful comments 👍⚖️👎. 🌊🌊🐳🌊🚣🌊🌊